\PassOptionsToPackage{unicode}{hyperref}
\PassOptionsToPackage{hyphens}{url}
\PassOptionsToPackage{dvipsnames,svgnames,x11names}{xcolor}
\documentclass[
  11pt,
]{article}
\usepackage{xcolor}
\usepackage[margin=1in]{geometry}
\usepackage{amsmath,amssymb}
\usepackage{iftex}
\ifPDFTeX
  \usepackage[T1]{fontenc}
  \usepackage[utf8]{inputenc}
  \usepackage{textcomp} 
\else 
  \usepackage{unicode-math} 
  \defaultfontfeatures{Scale=MatchLowercase}
  \defaultfontfeatures[\rmfamily]{Ligatures=TeX,Scale=1}
\fi
\usepackage{lmodern}
\ifPDFTeX\else
\fi
\IfFileExists{upquote.sty}{\usepackage{upquote}}{}
\IfFileExists{microtype.sty}{
  \usepackage[]{microtype}
  \UseMicrotypeSet[protrusion]{basicmath} 
}{}
\makeatletter
\@ifundefined{KOMAClassName}{
  \IfFileExists{parskip.sty}{%
    \usepackage{parskip}
  }{
    \setlength{\parindent}{0pt}
    \setlength{\parskip}{6pt plus 2pt minus 1pt}}
}{
  \KOMAoptions{parskip=half}}
\makeatother
\usepackage{longtable,booktabs,array}
\usepackage{caption}
\usepackage{calc} 
\usepackage{etoolbox}
\makeatletter
\patchcmd\longtable{\par}{\if@noskipsec\mbox{}\fi\par}{}{}
\makeatother
\usepackage{graphicx}
\makeatletter
\newsavebox\pandoc@box
\newcommand*\pandocbounded[1]{
  \sbox\pandoc@box{#1}%
  \Gscale@div\@tempa{\textheight}{\dimexpr\ht\pandoc@box+\dp\pandoc@box\relax}%
  \Gscale@div\@tempb{\linewidth}{\wd\pandoc@box}%
  \ifdim\@tempb\p@<\@tempa\p@\let\@tempa\@tempb\fi
  \ifdim\@tempa\p@<\p@\scalebox{\@tempa}{\usebox\pandoc@box}%
  \else\usebox{\pandoc@box}%
  \fi%
}
\def\fps@figure{htbp}
\makeatother
\providecommand{\tightlist}{%
  \setlength{\itemsep}{0pt}\setlength{\parskip}{0pt}}
\usepackage[]{natbib}
\usepackage{needspace}

\ifdefined\bibsep\fi
\usepackage{bookmark}
\IfFileExists{xurl.sty}{\usepackage{xurl}}{} 
\makeatletter
\@ifundefined{xmpquote}{}{}
\makeatother
\hypersetup{
  pdftitle={Easy to anticipate{,} hard to compute: boundary dependence finds the computed outputs that entropy patching misses},
  pdfauthor={Nicolás Vera Zúñiga --- Independent researcher, Chile --- nicovera@quetru.cl},
  colorlinks=true,
  linkcolor={blue},
  filecolor={Maroon},
  citecolor={blue},
  urlcolor={blue},
  pdfcreator={LaTeX via pandoc}}

\title{Easy to anticipate, hard to compute: boundary dependence finds
the computed outputs that entropy patching misses}
\author{Nicol\'as Vera Z\'u\~niga\\
Independent Researcher, Chile\\
\texttt{nicovera@quetru.cl}}
\date{}
\date{}

\ifPDFTeX
\DeclareUnicodeCharacter{2265}{\ensuremath{\geq}}
\DeclareUnicodeCharacter{2264}{\ensuremath{\leq}}
\DeclareUnicodeCharacter{00D7}{\ensuremath{\times}}
\DeclareUnicodeCharacter{2192}{\ensuremath{\rightarrow}}
\DeclareUnicodeCharacter{00B5}{\ensuremath{\mu}}
\DeclareUnicodeCharacter{00B0}{\ensuremath{^{\circ}}}
\DeclareUnicodeCharacter{00B1}{\ensuremath{\pm}}
\DeclareUnicodeCharacter{2212}{\ensuremath{-}}
\fi
\begin{document}
\maketitle

\begin{abstract}
Byte-level language models such as the Byte Latent Transformer (BLT)
group bytes into patches and run their large global model once per
patch. BLT starts a patch where a small model's next-byte entropy is
high. We show that under tight patch budgets this rule has a blind spot:
positions whose \emph{type} the preceding text fixes but whose
\emph{value} must be computed, such as the number after \texttt{=} in a
worked math solution. A nearly certain digit keeps the next byte's
entropy near or below \(\log_2 10 = 3.32\) bits, about BLT-1B's
threshold for a 10\% budget. In Meta's BLT-1B with patch starts on 10\%
of bytes, the entropy rule puts a patch start at 16\% of the computed
results in GSM8K solutions and gets 19.0\% of them exactly right given
the true preceding text; a boundary after each \texttt{=} at the same
patch count gets 51.8\%, and entropy combined with a label-free
\emph{boundary-dependence} signal gets 67.1\% (default layout at 26\% of
bytes: 76.8\%). The gap survives adapting BLT-1B to the budget with
low-rank adapters (32.9\% vs 72.7\%, three runs per rule). In byte
models trained from scratch at a 10\% budget on a 1.9 GB math corpus, it
is not detectable at 1M parameters, appears on final answers at 12M and
on both targets at 50M, as training data grow too: at 50M boundary
dependence gets 35.9\% of computed results against 13.9\% for entropy
and 78.9\% of final answers against 59.0\% (3 seeds each), although
final answers mostly copy the solution's last number and entropy starts
a patch at every one. In BLT-1B the blind spot narrows at larger budgets
(entropy covers about 90\% of the values right after \texttt{=} at 20\%
of bytes) and in the same solutions rewritten as prose, where a patch
start at each value still pays. Boundary dependence, the change in the
model's own loss when a patch start is removed or added, measured per
two-byte context on text in the same format, finds these positions
without labels.
\end{abstract}

\section{1. Introduction}\label{introduction}

Byte-level models avoid a fixed tokenizer by working on raw bytes, and
recover efficiency by grouping bytes into variable-length patches
\citep{pagnoni2025blt, slagle2024spacebyte, hwang2025hnet}. In BLT, a
small byte-level language model estimates the entropy of each next byte;
a patch starts where that entropy crosses a threshold, and the large
global transformer runs once per patch. The intuition is that compute
should go where prediction is hard.

Entropy measures uncertainty about the next byte. What a patch start
actually provides is different: a fresh global state for the bytes that
follow. These coincide in prose, where a surprising byte often begins a
new word. They come apart at a computed result. After
\texttt{16\ -\ 3\ -\ 4\ =}, the next byte is almost certainly a digit,
so its entropy stays near or below \(\log_2 10 = 3.32\) bits however
unsure the model is about which digit (BLT-1B's patcher is unsure there,
at a median of 2.8--3.0 bits). At a 10\% budget BLT-1B's entropy
threshold sits about that high, so it spends patches on hard-to-predict
prose and skips most of these positions (§5).

We make four contributions:

\begin{enumerate}
\def\labelenumi{\arabic{enumi}.}
\tightlist
\item
  \textbf{A blind spot in a trained 1B model.} At a 10\% patch budget,
  BLT-1B's own entropy rule skips most computed results in GSM8K
  solutions, and about half at 15\%; forcing a boundary after each
  \texttt{=} at the same patch count raises their exact match, given the
  true preceding text, by 19--33 points (§4). The cue before a result
  caps its entropy near BLT-1B's threshold at a 10\% budget; larger
  budgets and prose without \texttt{=} narrow the blind spot (§5).
\item
  \textbf{The gap survives training and adaptation.} BLT-style models
  trained from scratch at 10--20\% budgets show the same ordering across
  3 seeds. Trained at 1M, 12M and 50M parameters on a 1.9 GB math
  corpus, with training data growing alongside, the label-free trigger's
  advantage over entropy is not detectable at 1M and appears at 12M (+10
  points on final answers) and 50M (+20 on final answers, which mostly
  copy the solution's last number, and +22 on computed results; §6).
  BLT-1B adapted to a 10\% budget with low-rank adapters keeps a
  40-point gap (§8).
\item
  \textbf{A label-free trigger.} \emph{Boundary dependence}, the change
  in the model's own loss when a patch start is removed or added,
  averaged per preceding two-byte context, finds computed results
  without labels or hand-written rules, given text in the same format to
  fit it; it learns the contexts that precede values (§7).
\item
  \textbf{Scope.} The effect needs a computed value and a model able to
  compute it; it is largest after a cue that fixes the value's type and,
  in BLT-1B, at tight budgets, while in our small models it persists at
  15--25\% (§5, §6). Copied names and values and logic lookups gain 1--3
  points from a patch start at 10\%, and values the model cannot compute
  show no detectable gain (−0.3 points, −1.1 to +0.5; §9). Scratchpad
  Patching's entropy trigger shows no detectable difference from random
  scratchpads on final answers (−0.3 points, −2.2 to +1.6; §10).
\end{enumerate}

\section{2. Background and related
work}\label{background-and-related-work}

\textbf{Patching rules.} BLT \citep{pagnoni2025blt} describes two
entropy rules: a global threshold, \(H(x_t) > \theta_g\), and an
approximate monotonic rule that starts a patch where entropy rises,
\(H(x_t) - H(x_{t-1}) > \theta_r\). Its final version compares both at
8B scale and defaults to the global threshold when not specified, as do
the released BLT-1B checkpoint and the official code
(\texttt{monotonicity:\ false}, threshold 1.335); we call that rule
\emph{entropy} and the monotonic rule \emph{jump}. BLT also notes that
an entropy model with a small receptive field can be stored as a lookup
table, and its Figure 3 illustrates patching with a small entropy model
that sees two bytes; our dependence rule (§7) is such a table with a
different target. Entropy spikes as boundaries go back to dynamic
pooling \citep{nawrot2023dynamic}. SpaceByte \citep{slagle2024spacebyte}
starts patches at word boundaries; H-Net \citep{hwang2025hnet}, AU-Net
\citep{videau2025aunet}, MrT5 \citep{kallini2024mrt5} and ByteSpan
\citep{goriely2025bytespan} learn or derive boundaries in other ways.
AU-Net reports GSM8K accuracy as a benchmark, rising with the number of
hierarchy stages; none of these papers examines where boundaries fall on
computed outputs or measures answer accuracy as a function of where
boundaries are placed.

\textbf{Compute between patches.} Scratchpad Patching
\citep{zheng2026scratchpad} decouples compute from patch size by
inserting transient global steps (``scratchpads'') inside patches,
triggered by an absolute next-byte entropy threshold, and names the
stale-context problem \emph{patch lag}. Its trigger ablation compares
strategies on validation bits per byte (code, natural-language and math
splits) and finds the entropy trigger best; its downstream evaluation
(code pass@1, multiple-choice understanding) includes no math answer
accuracy. Fast BLT \citep{kallini2026fastblt} speeds up decoding without
changing where boundaries go. EntropyMoE \citep{liu2026entropymoe} uses
patch entropy to route each patch of a BLT-style model to feed-forward
experts; it is evaluated on bits per byte and downstream accuracy, not
on math.

\textbf{Tokenization and arithmetic.} For static tokenizers, how numbers
are segmented affects arithmetic
\citep{singh2024tokenization, meister2026tokeval}. Our setting differs:
boundaries are decided online by a model, and the question is which
positions receive a fresh global step.

\textbf{Selection signals.} Rho-1 \citep{lin2024rho1} selects training
tokens by excess loss against a reference model; we test a patch rule
built on the same signal as a baseline (§12). Mixture-of-Depths
\citep{raposo2024mod} routes compute per token by a learned router.

\section{3. Setup}\label{setup}

\textbf{BLT-1B.} We use the released BLT-1B through its Hugging Face
conversion (\texttt{itazap/blt-1b-hf}), which accepts explicit patch
lengths, so layouts can be changed at inference without retraining. The
conversion drops BLT's 512-byte sliding window in the entropy patcher
and in the local encoder and decoder (transformers issue 49185; it and
the fix proposed in pull request 49188 were still open on 1 November
2026). We restore the window with that fix (applied to transformers
5.18; the patch is in our repository) and use the corrected model
throughout, except where §13 reports the released conversion. Positions
before byte 512 are unchanged by the fix, since the window covers every
earlier byte; past it the released conversion over-segments, and 121 of
our 294 test problems are longer than 512 bytes. With the window, the
default layout starts a patch on 25.5\% of bytes on GSM8K, against
28.4\% without it. We build tight-budget layouts at R = 10\% and 15\% of
bytes. Thresholds are fitted once on the first 300 GSM8K training
problems (297 fit the context) so that the mean patch rate there is R,
then applied position by position at test time; whether a byte starts a
patch never depends on later bytes. (Choosing the top R\% within each
test problem instead gives the same conclusions; §13.)

\textbf{Layouts.} \emph{entropy}: BLT-1B's next-byte entropy.
\emph{results} (hand-written): a boundary right after every \texttt{=},
the rest filled by entropy. \emph{dep}: boundary dependence (§7).
\emph{entdep}: the sum of standardized entropy and dependence.

\textbf{Evaluation.} 294 GSM8K test problems (question, worked solution
with calculator annotations removed, and ``The final answer is N''),
containing 796 in-line computed results: the number right after
\texttt{=}, with its \texttt{\$} if any (181 are written
\texttt{=\ \$N}). A target counts as exactly right when every byte is
the model's top prediction given the true preceding text (teacher
forcing), and as covered when a patch starts at its first byte. We
report final answers as well. In GSM8K the final answer repeats the
solution's last number in 94.6\% of the 1,319 test problems (95.6\% of
our 294; thousands separators ignored), and that number ends a median of
8 bytes before the final-answer line, so under teacher forcing a final
answer is mostly a copy. We test paired differences with exact McNemar
tests and 95\% intervals that resample whole problems.

\textbf{Small models trained from scratch.} A BLT-style model in MLX: a
local encoder (byte embedding plus in-patch offset embedding),
cross-attention pooling with the patch mean as query, a global
transformer (D = 128, 4 layers unless noted; 1.1M parameters in all,
0.8M of them global), and a local decoder layer that attends to the
previous 32 bytes across patch boundaries. Patch starts come from a
fixed rule applied to the bytes and to a small entropy model (one layer,
32 dimensions) trained on the same corpus. Training: GSM8K and MATH
training solutions \citep{cobbe2021gsm8k, hendrycks2021math} (9.7 MB,
GSM8K's calculator annotations kept), 32,000 steps of 32 windows of 128
bytes, AdamW with warmup and cosine decay. Evaluation: 660 GSM8K and 700
MATH test problems held out from training, with greedy exact match given
the true prefix on 1,000 computed results (the number after \texttt{=}
inside an annotation such as
\texttt{\textless{}\textless{}16-3-4=9\textgreater{}\textgreater{}9},
copied after \texttt{\textgreater{}\textgreater{}}), 660 final answers
(after \texttt{\#\#\#\#}) and 700 MATH
\texttt{\textbackslash{}boxed\{\}} answers; patch boundaries during
decoding are decided online by the same rule. Three seeds per
configuration; rules are compared by the difference of seed means with a
Welch 95\% interval over seeds (degrees of freedom rounded down, which
widens it).

\section{4. The blind spot in BLT-1B}\label{the-blind-spot-in-blt-1b}

At BLT-1B's default budget its patcher covers computed results well
(79\% of them start a patch), and removing the patch start at the final
answer costs 19 points of exact match (97.0\% \(\rightarrow\) 78.3\%),
confirming that the boundary matters to BLT-1B, whose own layout starts
a patch there. (That test scores the final answer as the solution last
wrote it, with any \texttt{\$} and thousands separators, on 300
problems; Table 1 scores the plain number after ``The final answer is''
on the 294 problems that fit the context, which gives 73.5\% for the
default layout.) Under a tight budget, entropy drops computed results
faster than other positions: at 15\% it covers 48\% of them, at 10\%
only 16\%.

\textbf{Table 1.} BLT-1B, patch layouts changed at inference
(train-fitted thresholds). Exact match on 796 in-line computed results
and 294 final answers.

{\def\LTcaptype{none} 
{\footnotesize\setlength{\tabcolsep}{3pt}
\begin{longtable}[]{@{}
  >{\raggedright\arraybackslash}p{(\linewidth - 8\tabcolsep) * \real{0.4808}}
  >{\raggedright\arraybackslash}p{(\linewidth - 8\tabcolsep) * \real{0.0962}}
  >{\raggedright\arraybackslash}p{(\linewidth - 8\tabcolsep) * \real{0.1346}}
  >{\raggedright\arraybackslash}p{(\linewidth - 8\tabcolsep) * \real{0.1538}}
  >{\raggedright\arraybackslash}p{(\linewidth - 8\tabcolsep) * \real{0.1346}}@{}}
\toprule\noalign{}
\begin{minipage}[b]{\linewidth}\raggedright
Layout
\end{minipage} & \begin{minipage}[b]{\linewidth}\raggedright
Patch rate
\end{minipage} & \begin{minipage}[b]{\linewidth}\raggedright
Results covered
\end{minipage} & \begin{minipage}[b]{\linewidth}\raggedright
Computed results
\end{minipage} & \begin{minipage}[b]{\linewidth}\raggedright
Final answers
\end{minipage} \\
\midrule\noalign{}
\endhead
\bottomrule\noalign{}
\endlastfoot
default (BLT-1B's own) & 0.255 & 79\% & 76.8\% & 73.5\% \\
entropy@15 & 0.157 & 48\% & 47.0\% & 65.0\% \\
results@15 (hand-written) & 0.157 & 100\% & 66.1\% & 67.0\% \\
dep@15 & 0.153 & 100\% & 77.3\% & 40.8\% \\
entdep@15 & 0.157 & 100\% & 76.5\% & 67.3\% \\
entropy@10 & 0.106 & 16\% & 19.0\% & 51.4\% \\
results@10 (hand-written) & 0.106 & 100\% & 51.8\% & 60.5\% \\
dep@10 & 0.100 & 100\% & 67.8\% & 4.8\% \\
entdep@10 & 0.105 & 100\% & 67.1\% & 62.9\% \\
\end{longtable}
}
}

At equal patch counts, a boundary after each \texttt{=} beats entropy on
computed results by +19.1 points at 15\% (95\% interval +15.8 to +22.4;
172 results right only under \emph{results} vs 20 only under
\emph{entropy}; McNemar \(p = 2 \times 10^{-31}\)) and +32.8 at 10\%
(\(p = 7 \times 10^{-69}\)). All layouts run at or slightly above budget
on the test problems (patch rates 0.157 and 0.106 for entropy and the
hand-written layout at 15\% and 10\%, 0.100 to 0.157 for the others).
Figure 1 shows a single solution (425 bytes) with the top 10\% of its
positions by each score: entropy places its patch starts at word
beginnings and misses all six computed results; entropy plus dependence
starts a patch right before each of them.

\begin{figure}
\centering
\includegraphics[width=0.75\linewidth,height=\textheight,keepaspectratio,alt={BLT-1B
patch starts in one GSM8K solution at a 10\% budget, with equal patch
counts for the two layouts. Bars mark patch starts and shading marks
in-line computed results. Entropy starts no patch at any of the six
results; entropy plus dependence starts one before
each.}]{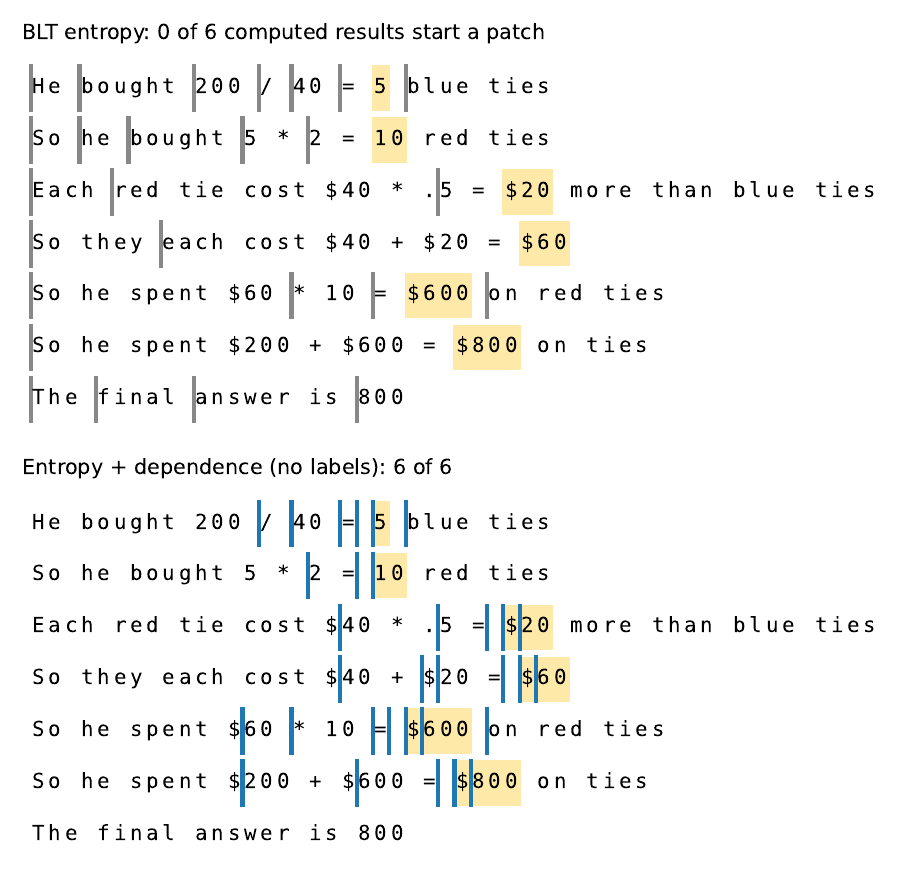}
\caption{BLT-1B patch starts in one GSM8K solution at a 10\% budget,
with equal patch counts for the two layouts. Bars mark patch starts and
shading marks in-line computed results. Entropy starts no patch at any
of the six results; entropy plus dependence starts one before each.}
\end{figure}

\section{5. What the blind spot depends
on}\label{what-the-blind-spot-depends-on}

Besides Table 1's 294 problems, this section uses 915 further GSM8K test
problems (those among problems 300--1,318 that fit the context and had a
valid prose rewrite, made for a measurement not reported here; 2,448
computed values) and a prose rewrite of 221 of the first 300 (711
values). On both, thresholds are fitted on the same 1,382 GSM8K training
problems in the writing tested, and a value counts as covered when a
patch starts at its first digit, after any \texttt{\$}. (Table 1 counts
the \texttt{\$}: there the default layout starts a patch at only 20\% of
the 181 results written \texttt{=\ \$N}, yet gets 71\% of them right,
against 78\% of the others.)

\textbf{The cue caps entropy.} If a digit is certain to come next, the
next byte's entropy cannot exceed \(\log_2 10 = 3.32\) bits, however
unsure the model is about which digit. BLT-1B's threshold for a 10\%
budget lies at about this level (3.36 bits), so after a cue that makes a
digit nearly certain, entropy rarely starts a patch. In Table 1's
results, entropy at 10\% starts a patch at none of the 181 written
\texttt{=\ \$N} and at 21\% of the 615 written \texttt{=\ N}. On the 915
problems the patcher gives a digit a median probability of 0.996 at the
first digit after \texttt{=\ \$} (median entropy 2.76 bits), where
entropy at 10\% starts no patch; after \texttt{=} the median is 0.966
(3.00 bits) and entropy covers 18.9\% of the values. Of all values it
misses at 10\%, 90.1\% (88.4 to 91.8) have a digit probability of at
least 0.9. Writing \texttt{is} instead of \texttt{=} before the 1,862
values written \texttt{=\ N} lowers that median from 0.966 to 0.736 and
raises entropy's coverage of them from 18.9\% to 65.7\% at the same
budget (+46.9 points, +44.5 to +49.3). We found the split by cue after
the fact, in the prose experiment below, and measured these problems
afterwards to test it.

\textbf{Budget.} In BLT-1B, as the budget grows the threshold falls
below the ceiling and the blind spot closes: on the 915 problems entropy
starts a patch at 14\% of computed values at 10\% of bytes, 56\% at
15\%, 89\% at 20\% and 97\% at 25\% (thresholds 3.36, 2.86, 2.38 and
1.90 bits), and after \texttt{=\ \$} at 0\%, 33\%, 85\% and 97\%. It
starts one at 94\% of final answers already at 10\%. BLT's own scaling
experiments use average patches of 6 and 8 bytes \citep{pagnoni2025blt},
about 17\% and 12.5\% of bytes; at 12.5\% entropy covers 28\% of
computed values. In our models trained from scratch, whose entropy model
has one layer of 32 dimensions, the gap persists at 15--25\% budgets
(§6).

\textbf{Format.} We also had an LLM rewrite the solutions as prose
without \texttt{=} or arithmetic symbols between numbers, keeping every
computed value in order with the same digits (checked automatically). At
10\%, entropy starts a patch at 57.2\% of the values in prose against
13.6\% in the original (+43.6 points, +38.8 to +48.2). Adding a start at
every value to entropy's layout, slightly beyond the budget, raises
exact match by 37.0 points in the original (+33.1 to +41.1) but only by
12.9 in prose (+10.3 to +15.7): the blind spot narrows but does not
disappear. Most of those 12.9 points come from the 163 values after
\texttt{\$}, a type-fixing cue the rewrite kept: entropy starts a patch
at none of them (nor at any of the original's 167), and a start adds 36
points there, against 6 after a plain space, where entropy already
covers 74\%. A start at every value still pays in prose: against the
same layout without those starts it adds 28.4 points (+24.7 to +32.2;
44.3 in the original). Values are harder in prose (default layout 56.8\%
against 77.9\%; BLT-1B may have seen the original solutions in
training), and prose has 1.43 times as many bytes per value, so the same
budget gives it more patch starts per value; the \texttt{is} swap above,
which changes only the cue, is the cleaner test.

\section{6. Training at tight budgets}\label{training-at-tight-budgets}

Changing layouts at inference puts BLT-1B out of distribution. To test
whether the blind spot survives a model that learned with the tight
layout, we train small byte models from scratch with each rule.

\textbf{Table 2.} Trained at a fixed budget (D = 128, 4 global layers,
1.1M parameters, 3 seeds). Computed results / final answers / bits per
byte. Every entropy seed is below every seed of the dependence and
hand-written rules on both accuracies; the hand-written and dependence
seeds overlap on computed results at 20\%.

{\def\LTcaptype{none} 
{\footnotesize\setlength{\tabcolsep}{3pt}
\begin{longtable}[]{@{}
  >{\raggedright\arraybackslash}p{(\linewidth - 8\tabcolsep) * \real{0.0682}}
  >{\raggedright\arraybackslash}p{(\linewidth - 8\tabcolsep) * \real{0.2273}}
  >{\raggedright\arraybackslash}p{(\linewidth - 8\tabcolsep) * \real{0.2273}}
  >{\raggedright\arraybackslash}p{(\linewidth - 8\tabcolsep) * \real{0.2386}}
  >{\raggedright\arraybackslash}p{(\linewidth - 8\tabcolsep) * \real{0.2386}}@{}}
\toprule\noalign{}
\begin{minipage}[b]{\linewidth}\raggedright
Budget
\end{minipage} & \begin{minipage}[b]{\linewidth}\raggedright
Entropy
\end{minipage} & \begin{minipage}[b]{\linewidth}\raggedright
Entropy jump (BLT's monotonic rule)
\end{minipage} & \begin{minipage}[b]{\linewidth}\raggedright
Dependence (label-free)
\end{minipage} & \begin{minipage}[b]{\linewidth}\raggedright
Hand-written results rule
\end{minipage} \\
\midrule\noalign{}
\endhead
\bottomrule\noalign{}
\endlastfoot
10\% & 7.1\% / 10.4\% / 1.654 & 7.0\% / 64.1\% / 1.639 & 17.1\% / 67.4\%
/ 1.678 & 24.7\% / 72.5\% / 1.640 \\
15\% & 8.3\% / 32.5\% / 1.552 & not run & 14.8\% / 68.2\% / 1.638 &
22.7\% / 74.8\% / 1.528 \\
20\% & 8.4\% / 44.0\% / 1.475 & not run & 17.4\% / 68.2\% / 1.577 &
23.9\% / 75.1\% / 1.451 \\
\end{longtable}
}
}

The ordering hand-written \textgreater{} dependence \textgreater{}
entropy holds at every budget in the seed means. BLT's own monotonic
rule (\emph{jump}) separates the two targets: at 10\% it recovers final
answers (64.1\% against 10.4\% for entropy) but not computed results
(7.0\% against 7.1\%; −0.1 points, −1.1 to +0.9); we did not establish
why. A patch start at the answer itself does not decide final answers
here: on held-out text at 10\%, neither entropy nor dependence starts a
patch at a final answer (they follow \texttt{\#\#\#\#}), yet dependence,
which starts one at 37\% of computed results (entropy: none) and at
every copy of a result after \texttt{\textgreater{}\textgreater{}}, gets
67.4\% of final answers. (Jump's recovery disappears on the larger
corpus below, where entropy itself starts a patch at every final answer,
so that comparison does not isolate model size.) Placement matters more
than budget: dependence at 10\% beats entropy at 20\% on final answers
by 23.4 points (95\% interval over seeds +17.2 to +29.5) with half the
patches. From 10\% to 20\%, entropy's final-answer accuracy rises (10\%
\(\rightarrow\) 33\% \(\rightarrow\) 44\%; +33.6 points, +21.2 to
+46.1), while the result-aware rules show no detectable change
(dependence +0.8, −4.3 to +5.9; hand-written +2.5, −3.5 to +8.6), nor
does any rule's computed-result accuracy (entropy +1.3, −2.4 to +5.0;
dependence +0.3, −8.5 to +9.2; hand-written −0.7, −13.2 to +11.8). MATH
answers stay near 3\% for every rule; we read this as arithmetic
capacity capping each rule's level at this size, while placement sets
the gaps between rules. The hand-written rule's bits per byte are lower
than dependence's at every budget and than entropy's at 15\% and 20\%,
and tied with entropy's and jump's at 10\% (−0.013, −0.046 to +0.019;
+0.002, −0.033 to +0.037), so its gain shows no detectable cost in bits
per byte; dependence costs 1.5--7\% in bits per byte against entropy.

The same pattern holds at entropy's usual 25\% budget (Figure 4). At D =
128, word starts plus math syntax (18.5\% of bytes) gets 30.4\% of
computed results against 11.0\% for entropy (+19.4 points, +11.0 to
+27.8, 3 seeds); word starts alone get 15.5\% and a 6-byte stride plus
math syntax 16.6\%, so word alignment and the result boundary each reach
only about half the combined accuracy. At D = 64 (0.2M parameters),
BLT's entropy rule gets 9.8\% of final answers, its jump variant 67.5\%,
and word starts plus math syntax 77.1\% (3 seeds each).

\textbf{Scaling to 50M parameters.} The models above are small (1.1M
parameters) and trained only on GSM8K and MATH solutions. To test
whether the gap shrinks as models grow, we trained the same architecture
from scratch at 1.1M, 11.7M and 53.7M parameters on a larger corpus: the
first four shards of OpenWebMath \citep{paster2024openwebmath} (221,264
documents, 1.71 GB, after dropping 324 that share a 13-word sequence
with a test problem), plus the GSM8K and MATH training solutions
repeated to 10\% of the bytes (1.90 GB in total). Each rule is applied
at 10\% of bytes with thresholds fitted on this corpus; the dependence
rule uses the trained-model table of §7 (fitted on the GSM8K and MATH
corpus), because a table refitted on this corpus is dominated by
web-text contexts and covers none of the computed-result starts. The
models see 0.13, 0.5 and 1.5 billion bytes. The code is a PyTorch port
of our MLX harness, checked against it (logits agree to 3e-6; masks are
identical), and the 28 runs of Table 3 ran on one rented A40 GPU for
about \$19.

\textbf{Table 3.} Models trained from scratch at a 10\% patch budget on
the scaling corpus (one A40 GPU, no gradient clipping). Exact match on
660 final answers and 1,000 computed results; patch rates are measured
on the evaluation text (thresholds are fitted on the training corpus, so
entropy spends slightly more than 10\% there).

{\def\LTcaptype{none} 
{\footnotesize\setlength{\tabcolsep}{3pt}
\begin{longtable}[]{@{}
  >{\raggedright\arraybackslash}p{(\linewidth - 12\tabcolsep) * \real{0.1075}}
  >{\raggedright\arraybackslash}p{(\linewidth - 12\tabcolsep) * \real{0.2151}}
  >{\raggedright\arraybackslash}p{(\linewidth - 12\tabcolsep) * \real{0.0538}}
  >{\raggedright\arraybackslash}p{(\linewidth - 12\tabcolsep) * \real{0.2581}}
  >{\raggedright\arraybackslash}p{(\linewidth - 12\tabcolsep) * \real{0.2581}}
  >{\raggedright\arraybackslash}p{(\linewidth - 12\tabcolsep) * \real{0.0538}}
  >{\raggedright\arraybackslash}p{(\linewidth - 12\tabcolsep) * \real{0.0538}}@{}}
\toprule\noalign{}
\begin{minipage}[b]{\linewidth}\raggedright
Parameters
\end{minipage} & \begin{minipage}[b]{\linewidth}\raggedright
Rule
\end{minipage} & \begin{minipage}[b]{\linewidth}\raggedright
Seeds
\end{minipage} & \begin{minipage}[b]{\linewidth}\raggedright
Final answers (by seed)
\end{minipage} & \begin{minipage}[b]{\linewidth}\raggedright
Computed results (by seed)
\end{minipage} & \begin{minipage}[b]{\linewidth}\raggedright
Bits per byte
\end{minipage} & \begin{minipage}[b]{\linewidth}\raggedright
Eval patch rate
\end{minipage} \\
\midrule\noalign{}
\endhead
\bottomrule\noalign{}
\endlastfoot
1.1M & entropy & 3 & 7.0\% (8.8, 4.8, 7.4) & 4.4\% (3.6, 4.9, 4.8) &
2.093 & 11.2\% \\
1.1M & dependence & 2 & 5.4\% (6.2, 4.5) & 4.2\% (4.3, 4.1) & 2.196 &
10.9\% \\
1.1M & hand-written & 3 & 45.5\% (54.8, 25.3, 56.2) & 4.6\% (3.9, 4.1,
5.9) & 2.109 & 12.1\% \\
11.7M & entropy & 3 & 46.6\% (42.0, 49.1, 48.8) & 5.8\% (4.8, 6.4, 6.3)
& 1.657 & 11.2\% \\
11.7M & dependence & 3 & 56.7\% (58.5, 53.9, 57.6) & 7.7\% (7.2, 5.6,
10.3) & 1.673 & 10.9\% \\
11.7M & hand-written & 3 & 76.3\% (76.4, 73.9, 78.5) & 12.9\% (14.9,
10.3, 13.5) & 1.645 & 12.1\% \\
53.7M & entropy & 3 & 59.0\% (58.8, 57.9, 60.5) & 13.9\% (16.4, 10.4,
14.9) & 1.439 & 11.2\% \\
53.7M & dependence & 3 & 78.9\% (78.5, 80.2, 78.0) & 35.9\% (32.7, 35.7,
39.4) & 1.420 & 10.9\% \\
53.7M & hand-written & 2 & 85.5\% (85.8, 85.3) & 43.4\% (36.8, 49.9) &
1.431 & 12.1\% \\
53.7M & jump (BLT monotonic) & 1 & 53.8\% (53.8) & 6.3\% (6.3) & 1.410 &
9.7\% \\
53.7M & entropy at 20\% & 1 & 80.2\% (80.2) & 23.1\% (23.1) & 1.146 &
22.8\% \\
\end{longtable}
}
}

\begin{figure}
\centering
\includegraphics[width=0.9\linewidth,height=\textheight,keepaspectratio,alt={Exact
match against parameter count for models trained from scratch at a 10\%
budget (small markers: seeds; large markers: means). The label-free
dependence rule's advantage over BLT's entropy rule is not detectable at
1.1M, appears on final answers at 11.7M and on both targets at 53.7M,
where parameters and training data grow together. BLT's jump rule and
entropy at twice the budget were run at 50M only.}]{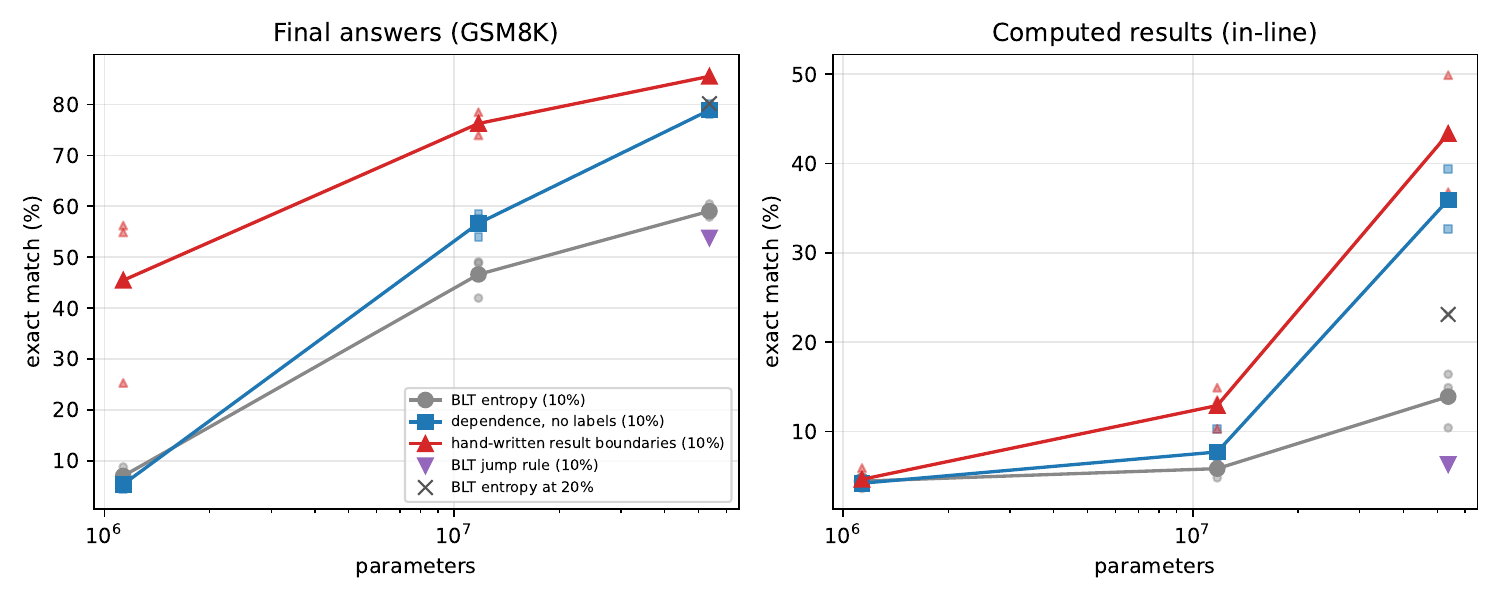}
\caption{Exact match against parameter count for models trained from
scratch at a 10\% budget (small markers: seeds; large markers: means).
The label-free dependence rule's advantage over BLT's entropy rule is
not detectable at 1.1M, appears on final answers at 11.7M and on both
targets at 53.7M, where parameters and training data grow together.
BLT's jump rule and entropy at twice the budget were run at 50M only.}
\end{figure}

Dependence is ahead of entropy on final answers at 11.7M and 53.7M and
on computed results at 53.7M (Figure 2), but parameters, training bytes
and passes over the solutions grow together. On final answers the
difference is −1.6 points at 1.1M (95\% interval over seeds −7.8 to
+4.5), +10.1 at 11.7M (+1.4 to +18.7) and +19.8 at 53.7M (+16.7 to
+23.0; 78.9\% against 59.0\%), with every dependence seed above every
entropy seed at the two larger sizes. These gaps do not come from patch
starts at the final answer: on the evaluation text, entropy at 10\%
starts a patch at every final answer and dependence at none. Final
answers mostly repeat the solution's last number (§3; 94.2\% of the 660
evaluation problems), and at 11.7M and 53.7M their accuracy is ordered
like the share of computed results that get a patch start (entropy
0.7\%, dependence 36.7\%, hand-written 100\%); we did not test whether
one causes the other.

On computed results, dependence and entropy separate only at 53.7M: at
1.1M and 11.7M they show no detectable difference (−0.2 points, −2.1 to
+1.6; +1.9, −4.5 to +8.2), and at 53.7M dependence gets 35.9\% against
13.9\% for entropy (+22.0 points, +13.6 to +30.4), again with every seed
above every entropy seed, and the hand-written rule 43.4\% (two seeds).
Dependence also has lower bits per byte than entropy (1.420 against
1.439; −0.019, −0.034 to −0.003). With a second recipe, gradient
clipping at norm 1.0 on the same GPU type (3 seeds per rule), dependence
again leads entropy at 53.7M on final answers, 78.2\% against 59.9\%
(+18.3 points, +15.5 to +21.1). On computed results every dependence
seed is again above every entropy seed, but the interval includes 0:
35.1\% against 20.7\% (+14.3, −5.7 to +34.4; entropy's seeds spread from
12.1\% to 27.5\%). Bits per byte show no detectable difference (+0.005,
−0.016 to +0.025).

BLT's jump rule does not help on this corpus: one seed gets 53.8\% of
final answers and 6.3\% of computed results, below every entropy seed,
although it has the lowest bits per byte of the rules in Table 3 at a
10\% budget (1.410). Given twice the budget (20\%, which is 22.8\% of
the evaluation text), entropy gets 80.2\% of final answers (one seed),
within the range of the dependence seeds at 10\% (78.0--80.2\%), and
23.1\% of computed results against 35.9\%: its extra patches cannot
reach the answer, which it already covers at 10\%, and its final answers
rise together with its computed results (13.9\% to 23.1\%). At 1.1M on
this corpus only the hand-written rule rises above the floor on final
answers; these models see the GSM8K and MATH solutions about 1.4 times,
against about 14 times for the models of Table 2, and their entropy
model covers every final answer, against none there, so either
difference could separate the two 1.1M results. One of the 28 runs of
Table 3 (1.1M, dependence, seed 0) never learned (bits per byte 4.72
against 2.08--2.26 for the other 1.1M runs) and is excluded.

\section{7. A label-free trigger: boundary
dependence}\label{a-label-free-trigger-boundary-dependence}

The hand-written rule uses domain knowledge (\texttt{=}). We look for a
signal that finds the same positions from the model alone.

\textbf{Definition.} For BLT-1B: on 400 GSM8K training problems, compute
the per-byte loss under the default layout and under a 10\% entropy
layout. For each position t, take the mean rise in loss over bytes
t\ldots t+3, and average it per preceding two-byte context
\((b_{t-2}, b_{t-1})\). At inference the score is a causal table lookup.
The top contexts are arithmetic (\texttt{=\$}, \texttt{0*}, \texttt{5=},
\texttt{3=}); after \texttt{=} the mean rise is 0.88 nats against 0.13
on average.

For the small trained models the same recipe fails: a reference model
trained with entropy patches rarely had a boundary at a result, so it
never learned to use one, and removing one costs little. Training the
reference model with random patch starts (25\%) and measuring the drop
in loss from \emph{adding} one patch start at t to a random 10\% layout
fixes this; the top contexts then include
\texttt{\textless{}\textless{}}, \texttt{=6}, \texttt{x=} and
\texttt{\textgreater{}\textgreater{}}.

\textbf{Results.} On BLT-1B (Table 1), entropy plus dependence beats
entropy on computed results by +29.5 points at 15\% (+25.6 to +33.4) and
+48.1 at 10\%, and beats even the hand-written rule by +10.4 (+7.7 to
+13.2): the table also places boundaries at operands and operators, not
only after \texttt{=}. On final answers it beats entropy by +11.6 points
at 10\% (+6.5 to +16.7; \(p = 3 \times 10^{-5}\)) but not at 15\% (+2.4,
−1.0 to +5.8), and it is tied with the hand-written rule (+2.4, −2.4 to
+6.8 at 10\%). Entropy starts a patch at nearly every final answer at
both budgets (93.5\% and 98.3\%); what differs is how many computed
results it covers (16\% and 48\%). Dependence alone fails on final
answers on BLT-1B (40.8\% and 4.8\%), because ``The final answer is'' is
not an arithmetic context; in the trained models dependence alone scores
higher on every seed (Table 2; entropy plus dependence: 9.7\% / 22.5\%,
2 seeds), though with two seeds the intervals include 0 (+7.4 points,
−7.4 to +22.1; +44.9, −74.6 to +164.4).

\section{8. Adapting BLT-1B to the
budget}\label{adapting-blt-1b-to-the-budget}

To test whether BLT-1B can learn its way around the blind spot, we train
low-rank adapters (rank 16, 16.9M parameters, on the global transformer
and the decoder's cross-attention) for 1,500 steps on GSM8K training
solutions under each 10\% layout, three runs per rule. Each model is
tested on the same 796 results under its own training layout.

\textbf{Table 4.} BLT-1B adapted to a 10\% budget, computed results
exact.

{\def\LTcaptype{none} 
{\footnotesize\setlength{\tabcolsep}{3pt}
\begin{longtable}[]{@{}llll@{}}
\toprule\noalign{}
Trained and tested under & Untrained & Runs & Mean \\
\midrule\noalign{}
\endhead
\bottomrule\noalign{}
\endlastfoot
entropy & 19.0\% & 32.8, 32.9, 33.0 & 32.9\% \\
results (hand-written) & 51.8\% & 62.4, 62.9, 62.7 & 62.7\% \\
entdep (label-free) & 67.1\% & 72.9, 70.9, 74.4 & 72.7\% \\
\end{longtable}
}
}

Adaptation helps every layout but closes none of the gap: entdep −
entropy is +39.8 points (pooled over runs, +36.3 to +43.2;
\(p = 6 \times 10^{-217}\)), hand-written − entropy +29.8, and entdep −
hand-written +10.0. The models trained under entropy do better under
entdep (66.2\% on average), a layout they never trained on, than under
their own (32.9\%). Final answers saturate near 97\% after fine-tuning,
because the final answer mostly repeats the solution's last number (§3)
and fine-tuning teaches the copy; they no longer separate the layouts.

\section{9. Which positions are
affected}\label{which-positions-are-affected}

If the blind spot is about computed values, copies and lookups should
gain little from a patch start. We repeat the BLT-1B test on four other
target types at 10\% (train-fitted thresholds).

\textbf{Table 5.} BLT-1B at 10\%, exact match. \emph{Forced}: a patch
start at every target, the rest by entropy.

{\def\LTcaptype{none} 
{\footnotesize\setlength{\tabcolsep}{3pt}
\begin{longtable}[]{@{}
  >{\raggedright\arraybackslash}p{(\linewidth - 10\tabcolsep) * \real{0.4494}}
  >{\raggedright\arraybackslash}p{(\linewidth - 10\tabcolsep) * \real{0.0787}}
  >{\raggedright\arraybackslash}p{(\linewidth - 10\tabcolsep) * \real{0.0787}}
  >{\raggedright\arraybackslash}p{(\linewidth - 10\tabcolsep) * \real{0.0674}}
  >{\raggedright\arraybackslash}p{(\linewidth - 10\tabcolsep) * \real{0.0674}}
  >{\raggedright\arraybackslash}p{(\linewidth - 10\tabcolsep) * \real{0.2584}}@{}}
\toprule\noalign{}
\begin{minipage}[b]{\linewidth}\raggedright
Target
\end{minipage} & \begin{minipage}[b]{\linewidth}\raggedright
Default
\end{minipage} & \begin{minipage}[b]{\linewidth}\raggedright
Entropy
\end{minipage} & \begin{minipage}[b]{\linewidth}\raggedright
Entdep
\end{minipage} & \begin{minipage}[b]{\linewidth}\raggedright
Forced
\end{minipage} & \begin{minipage}[b]{\linewidth}\raggedright
Kind
\end{minipage} \\
\midrule\noalign{}
\endhead
\bottomrule\noalign{}
\endlastfoot
GSM8K computed results & 76.8\% & 19.0\% & 67.1\% & 51.8\% & computed,
skill present \\
Python identifiers repeating a nearby name & 83.3\% & 81.7\% & 81.3\% &
83.1\% & copy \\
Proof-step conclusions, generated logic & 69.1\% & 46.3\% & 41.5\% &
49.4\% & rule lookup \\
Copied values, generated program traces & 82.5\% & 58.3\% & 62.3\% &
61.0\% & copy \\
Computed values, program traces & 20.5\% & 10.0\% & 9.0\% & 9.7\% &
computed, skill absent \\
\end{longtable}
}
}

These copies and lookups gain 1--3 points from a patch start (code:
+1.4, \(p = 4 \times 10^{-5}\)), against 32.8 for computed results. This
does not hold for copying in general: BLT-1B's final answers, mostly
copies of the solution's last number, need a start at the answer (§4).
Where BLT-1B cannot compute the value (program-trace arithmetic, 20.5\%
even with default patches) or answer at all (logic Yes/No answers at
chance under every layout), placement has nothing to act on. Small
models trained on the logic and trace tasks give the same picture: logic
is not learned (answers at chance), and on traces, where each value's
operands sit inside the local decoder's 32-byte window, entropy skips
most values (83\% at 10\%), yet the hand-written rule is tied with it
(21.8\% vs 20.6\%, 2 seeds; +1.2 points, 95\% interval over seeds −23.1
to +25.5). With the window cut to 8 bytes the hand-written rule is 4.1
points higher (24.2\% vs 20.1\%, 3 seeds), with no detectable difference
(−8.4 to +16.5). Our reading is that a start pays when a model able to
compute the value computes it, possibly only when the operands lie
beyond the local model's reach (the window test is inconclusive); type
and task are also confounded in Table 5.

\section{10. Scratchpad Patching}\label{scratchpad-patching}

We reimplemented Scratchpad Patching in the small-model harness (16-byte
fixed patches, scratchpads on 6\% of bytes, pooled by cross-attention
over the partial patch with the mean as query) and compared triggers at
equal scratchpad counts (D = 64, 2 global layers, 0.2M parameters).

\textbf{Table 6.} Final-answer exact match on GSM8K by scratchpad
trigger, with the accuracy of each seed.

{\def\LTcaptype{none} 
{\footnotesize\setlength{\tabcolsep}{3pt}
\begin{longtable}[]{@{}
  >{\raggedright\arraybackslash}p{(\linewidth - 6\tabcolsep) * \real{0.5063}}
  >{\raggedright\arraybackslash}p{(\linewidth - 6\tabcolsep) * \real{0.0633}}
  >{\raggedright\arraybackslash}p{(\linewidth - 6\tabcolsep) * \real{0.0886}}
  >{\raggedright\arraybackslash}p{(\linewidth - 6\tabcolsep) * \real{0.3418}}@{}}
\toprule\noalign{}
\begin{minipage}[b]{\linewidth}\raggedright
Trigger
\end{minipage} & \begin{minipage}[b]{\linewidth}\raggedright
Seeds
\end{minipage} & \begin{minipage}[b]{\linewidth}\raggedright
Final answers
\end{minipage} & \begin{minipage}[b]{\linewidth}\raggedright
By seed
\end{minipage} \\
\midrule\noalign{}
\endhead
\bottomrule\noalign{}
\endlastfoot
answer starts (math syntax) & 5 & 38.1\% & 49.2, 31.4, 8.3, 49.2,
52.4 \\
entropy (the paper's trigger) & 5 & 5.6\% & 7.3, 4.8, 4.2, 5.6, 6.1 \\
random positions & 5 & 5.9\% & 6.1, 5.6, 4.5, 5.2, 8.0 \\
none & 3 & 6.1\% & 6.2, 6.8, 5.2 \\
denser fixed patches, same compute (8-byte) & 5 & 7.8\% & 6.7, 12.3,
7.6, 5.2, 7.1 \\
\end{longtable}
}
}

There is no detectable difference between the entropy trigger and random
scratchpads (5.6\% vs 5.9\%, 5 seeds; −0.3 points, Welch 95\% interval
−2.2 to +1.6), while answer-start scratchpads give 38.1\%, more than six
times as much; one of their seeds stays at 8.3\%, with the highest bits
per byte of the five (1.922, against 1.900--1.909), in the range of the
entropy and random triggers (1.911--1.930). Computed results stay near
the floor under every trigger (4.7--6.3\%), so this test bears on final
answers, not on computed values. It does not contradict the paper's
ablation, which found entropy best on bits per byte: averaged bits per
byte can hide a failure at a few positions that decide answer accuracy.

\section{11. Compute}\label{compute}

Fewer patches save global compute, but local layers run on every byte.
For BLT-1B, counting matrix-multiply FLOPs, the local layers cost 299
MFLOPs per byte, the global model 2,653 MFLOPs per patch, and the
entropy model 199 MFLOPs per byte. A 10\% budget then costs 65\% of the
default's forward compute and 15\% costs 76\%. A dependence-table rule
is a lookup and needs no entropy model: 48\% at 10\% and 59\% at 15\%
(an entropy model stored as a lookup table, as BLT suggests, would also
avoid that cost). On computed results, dependence alone at 15\% is tied
with the default layout (77.3\% against 76.8\%; +0.5 points, −1.7 to
+2.7) at about 60\% of its compute (Figure 3), but it loses 32.7 points
on final answers (−38.1 to −27.2; §7). In our small models the local
layers dominate, so a 10\% budget saves only 24\% of compute (Figure 4).

\begin{figure}
\centering
\includegraphics[width=0.9\linewidth,height=\textheight,keepaspectratio,alt={BLT-1B
exact match on in-line computed results (left) and final answers (right)
against forward compute per byte, relative to the default layout, for
layouts at 10\% and 15\% of bytes. Dependence alone is a lookup and pays
no entropy-model cost.}]{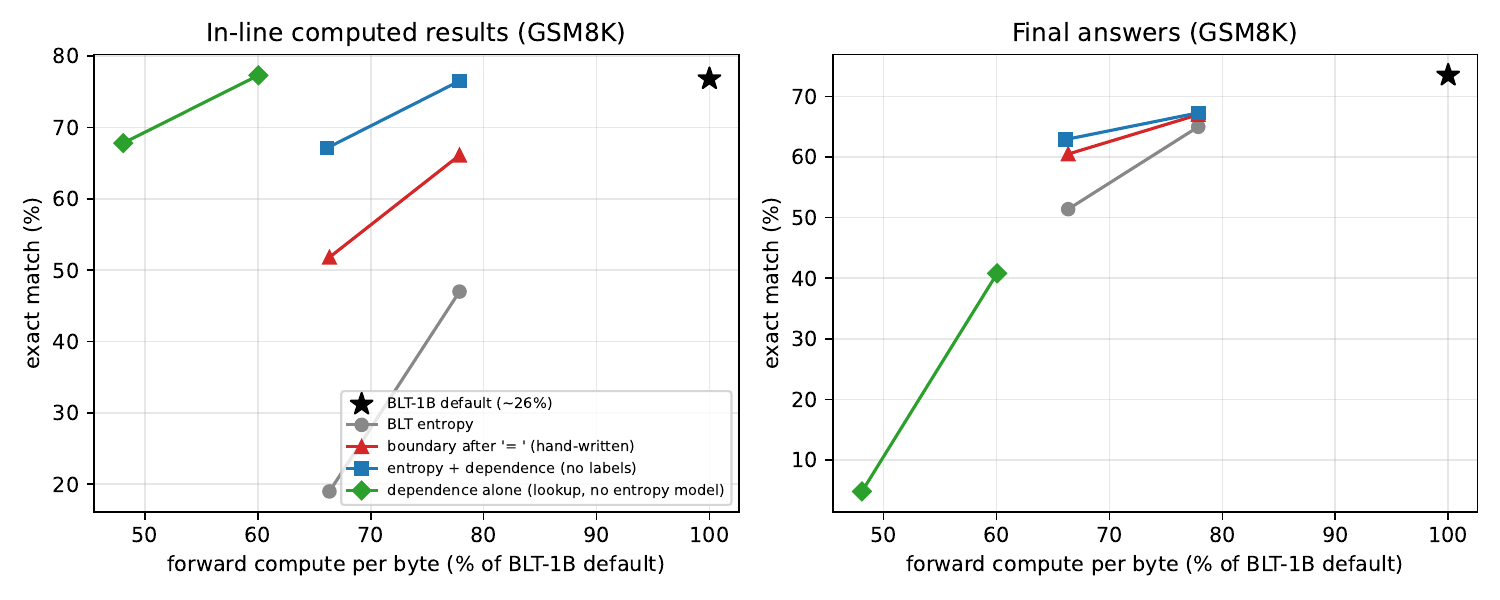}
\caption{BLT-1B exact match on in-line computed results (left) and final
answers (right) against forward compute per byte, relative to the
default layout, for layouts at 10\% and 15\% of bytes. Dependence alone
is a lookup and pays no entropy-model cost.}
\end{figure}

\begin{figure}
\centering
\includegraphics[width=0.9\linewidth,height=\textheight,keepaspectratio,alt={Models
trained at 10-25\% patch budgets (D = 128): exact match against forward
compute per byte, relative to entropy at 25\%. Points are seed means;
bars span the seeds.}]{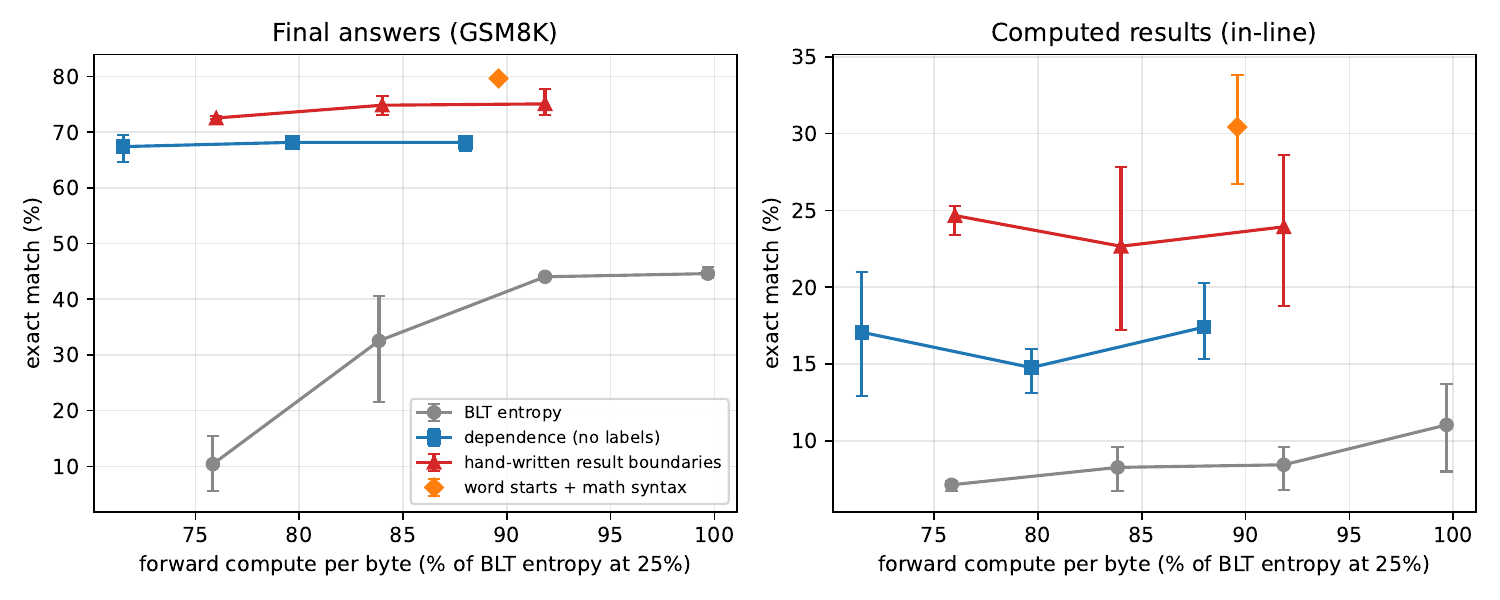}
\caption{Models trained at 10-25\% patch budgets (D = 128): exact match
against forward compute per byte, relative to entropy at 25\%. Points
are seed means; bars span the seeds.}
\end{figure}

\section{12. Negative results}\label{negative-results}

\textbf{Spending patches on hard or learnable bytes does not help.} A
Rho-1-style rule (patches where a small model's loss exceeds a reference
model's) and simplified learned chunking lost to hand-placed boundaries
in every setting tried.

\textbf{Finer arithmetic boundaries do not help.} Per-digit patches
after \texttt{=} and operand-aligned patches looked promising at two
seeds (on computed results and final answers); at five seeds neither
differs detectably from a matched random control (+0.7 points, −1.2 to
+2.6; +1.9, −2.9 to +6.7).

\textbf{A learned patcher does not beat the table.} A small network that
predicts the dependence gain from the last 8 bytes, trained on math and
on logic/trace data together, has held-out correlations close to each
domain's own table (math 0.140 against 0.140; logic and traces 0.147
against 0.158) and covers 83\% of computed-result starts (table: 37\%),
but trained into a model at 10\% it gets 14.3\% / 57.4\% against the
table's 17.1\% / 67.4\% (−2.7 points, −15.9 to +10.4; −10.0, −18.2 to
−1.8).

\textbf{Token models are partly protected.} With a Qwen2.5-0.5B draft
and a 1.5B target on generated program traces, the draft agrees with the
target on only 53\% of computed-value tokens (86\% on text), so
speculative decoding loses speed there; but the draft is rarely
confident on them (6.5\%), so a confidence-based early exit would let
few errors through. On GSM8K, computed results are the easiest tokens
for these models; we did not test why (overlap with their training data
is one possibility). In token models a computed value's uncertainty
shows in the model's own confidence; in byte patching, the decision to
start a patch before a result is made from bytes whose type is
predictable.

\textbf{The original synthetic result was mostly architectural.} In an
earlier synthetic benchmark with sum pooling and a patch-local decoder,
splitting a computation across patches cost up to a full bit; with
cross-attention pooling and a decoder that sees across patch boundaries,
most of that penalty disappeared. All results above use the
cross-attention model.

\section{13. Limitations}\label{limitations}

\textbf{Scale.} Models trained from scratch reach 53.7M parameters, on
1.5 billion bytes (the models of Table 2 have 1.1M; D = 64 is 0.2M).
Parameters, training bytes and passes over the GSM8K and MATH solutions
(about 1.4 at 1.1M, about 15 at 53.7M) grow together, three sizes on one
corpus show a trend, not a law, and at 53.7M the gap on computed results
is smaller with gradient clipping, and its interval then includes 0
(+14.3, −5.7 to +34.4, against +22.0, +13.6 to +30.4 without). The
models are small in every part: a 128-byte context, one local decoder
layer over 32 bytes and a one-layer, 32-dimensional entropy model.
BLT-1B results come from inference-time layout changes and low-rank
adaptation, not from full training at a tight budget. At 50M the
hand-written rule has two seeds and the jump and double-budget results
one each.

\textbf{Teacher forcing.} All accuracies score each target given the
true preceding text, so a final answer is mostly a copy of the
solution's last number (§3). End to end, BLT-1B cannot solve GSM8K by
itself: even after adaptation it gets about 2\% of final answers under
every layout (measured with the released conversion), so the gaps above
could not be measured end to end. The models trained from scratch also
solve essentially no GSM8K problems end to end at any size (at most 1 of
100 for any run).

\textbf{Domain.} The effect is established for arithmetic in worked math
solutions written with cues that fix the next byte's type (\texttt{=},
\texttt{\$}), in BLT-1B at budgets of 10--15\% of bytes (§5) and in
models trained from scratch at 10--25\% (§6); the models trained from
scratch saw only GSM8K's annotated format. Code shows a small,
significant effect; logic and program traces were inconclusive: the
models lacked the skill, or the local model may already have seen the
operands (§9).

\textbf{Label-free trigger.} The dependence table is fitted on the
domain it is tested on and does not transfer across formats (the GSM8K
table covers 0\% of program-trace values; a table refitted on the
scaling corpus covers no computed-result start, §6). It learns the
contexts that precede values, so it inherits their format. On BLT-1B,
the best variant (entropy plus dependence) differs from the one that
scores higher on every seed in trained models (dependence alone; §7),
which does not reach the hand-written rule. A follow-up
\citep{verazuniga2026patch} used another two-byte table, fitted without
labels to predict repeats of recent text, which on GSM8K works as a
learned detector of the cue before values; its version 2 corrects the
main claims of its first version.

\textbf{Checkpoint conversion.} The public BLT-1B conversion lacks the
512-byte sliding window; our BLT-1B results use it with the window
restored by the fix proposed upstream (§3). With the released conversion
the conclusions are the same, with different magnitudes (results −
entropy at 15\% and 10\%: +25.0 and +26.6 points).

\textbf{Thresholds.} Our original BLT-1B layouts chose the top R\% of
positions within each test problem, so a position's selection could
depend on later bytes. Train-fitted, position-by-position thresholds
reproduce every conclusion, with similar or larger gaps (released
conversion, results − entropy: +25.0 against +21.5 points at 15\%, +26.6
against +27.1 at 10\%).

\section{14. Conclusion}\label{conclusion}

Entropy tells a byte-level model where the next byte is surprising, not
where its decoder needs the global model. The two diverge at computed
outputs whose type the preceding text fixes: once only a digit can
follow, the next byte's entropy stays near or below \(\log_2 10\) bits,
about BLT-1B's threshold at a 10\% budget. Under tight budgets,
entropy-triggered patching skips these positions and teacher-forced
accuracy on them collapses: in a trained 1B model, after adaptation, and
in models trained at the budget (1.1M parameters on the solutions alone,
53.7M on the larger corpus). In BLT-1B, larger budgets and cues that do
not fix the type narrow the blind spot. Boundary dependence, a two-byte
table fitted without labels on text in the same format, finds many of
these positions by learning the contexts that precede them. Where such
cues precede values and budgets are tight, patch starts should follow
what the decoder needs from the global model rather than how surprising
the next byte is.

\section{Changes in version 2}\label{changes-in-version-2}

Version 1's experiments and the data in its tables and figures are
unchanged; version 2 adds later measurements that bound or correct its
claims (§5, §6), and version 1's §5--§13 are now §6--§14. The blind spot
is stated for type-fixing cues and, in BLT-1B, tight budgets (§5). Final
answers are described as mostly copies under teacher forcing (§3), and
explanations that assumed a patch start at the answer are corrected
(§6--§8). Version 1's statements that the gap grows with model size
(size and training data grew together) and that Scratchpad Patching's
entropy trigger shows the same blind spot are no longer made (§6, §10);
the reading of §9 and the thresholds comparison (§13) are restated. The
scaling study adds intervals over seeds and a second training recipe
(§6). New and corrected comparisons carry intervals; a p-value (§4), a
bits-per-byte ranking (§6) and Appendix A's seed counts are corrected.
§13 cites a follow-up, arXiv:2610.22100 \citep{verazuniga2026patch},
whose version 2 corrects the main claims of its first version. New
scripts, results and data, with the LLM-written rewrites of §5, are in
\texttt{paper/v2/}.

\section{Declarations}\label{declarations}

\textbf{Data and code.} Code, raw logs and the results registry are at
\url{https://github.com/nicoveraz/segresearch} (release v2.1.0 for this
version), archived at \url{https://doi.org/10.5281/zenodo.23238215}.
Scripts: \texttt{realblt\_budget.py}, \texttt{realblt\_code.py},
\texttt{realblt\_reason.py} and \texttt{realblt\_finetune.py} run the
BLT-1B experiments (PyTorch); \texttt{mathexp.py},
\texttt{deptrigger.py} and \texttt{reasonexp.py} the trained models
(MLX); \texttt{scale/} the scaling study (PyTorch, with checks against
the MLX code and the scripts that ran it on a rented GPU);
\texttt{flops.py} the compute estimates; \texttt{stats.py},
\texttt{make\_tables.py} and \texttt{make\_figures.py} rebuild the
statistics and the other tables and figures from \texttt{results/}, and
\texttt{paper/v2/scripts/scaling\_unclipped.py} rebuilds Table 3 and
Figure 2 from the runs in \texttt{results/scale/} without gradient
clipping. \texttt{paper/v2/} holds the scripts, results and data new in
version 2 (for §5's 915 problems, the measured rows but not the
measuring script). GSM8K, MATH and the BLT-1B weights are available from
their authors. \textbf{Funding:} none. \textbf{Competing interests:}
none. \textbf{Use of AI:} the experiments were designed, run and
analysed with an AI coding agent (Claude Code) working in a research
loop under the author's direction; the standards in Appendix A came out
of that process. The prose rewrites of §5 were written by an LLM (Claude
Opus 5.5) from one prompt and checked automatically; the rewrites and
the prompt are in \texttt{paper/v2/}. \textbf{Ethics:} public datasets
and generated text only; no human subjects.

\bibliographystyle{plainnat}
\bibliography{refs}

\begin{thebibliography}{18}
\providecommand{\natexlab}[1]{#1}
\providecommand{\url}[1]{\texttt{#1}}
\expandafter\ifx\csname urlstyle\endcsname\relax
  \providecommand{\doi}[1]{doi: #1}\else
  \providecommand{\doi}{doi: \begingroup \urlstyle{rm}\Url}\fi

\bibitem[Cobbe et~al.(2021)]{cobbe2021gsm8k}
Karl Cobbe et~al.
\newblock Training verifiers to solve math word problems.
\newblock \emph{arXiv preprint arXiv:2110.14168}, 2021.

\bibitem[Goriely et~al.(2025)Goriely, Salhan, Lesci, Cheng, and
  Buttery]{goriely2025bytespan}
Z{\'e}bulon Goriely, Suchir Salhan, Pietro Lesci, Julius Cheng, and Paula
  Buttery.
\newblock {ByteSpan}: Information-driven subword tokenisation.
\newblock In \emph{TokShop 2025 (non-archival)}, 2025.
\newblock arXiv:2506.18639.

\bibitem[Hendrycks et~al.(2021)]{hendrycks2021math}
Dan Hendrycks et~al.
\newblock Measuring mathematical problem solving with the {MATH} dataset.
\newblock \emph{arXiv preprint arXiv:2103.03874}, 2021.

\bibitem[Hwang et~al.(2026)Hwang, Wang, and Gu]{hwang2025hnet}
Sukjun Hwang, Brandon Wang, and Albert Gu.
\newblock Dynamic chunking for end-to-end hierarchical sequence modeling.
\newblock In \emph{International Conference on Learning Representations}, 2026.
\newblock arXiv:2507.07955. Code and checkpoints:
  https://github.com/goombalab/hnet.

\bibitem[Kallini et~al.(2024)Kallini, Murty, Manning, Potts, and
  Csord{\'a}s]{kallini2024mrt5}
Julie Kallini, Shikhar Murty, Christopher~D. Manning, Christopher Potts, and
  R{\'o}bert Csord{\'a}s.
\newblock {MrT5}: Dynamic token merging for efficient byte-level language
  models.
\newblock \emph{arXiv preprint arXiv:2410.20771}, 2024.

\bibitem[Kallini et~al.(2026)Kallini, Pagnoni, Limisiewicz, Ghosh, Zettlemoyer,
  Potts, Han, and Iyer]{kallini2026fastblt}
Julie Kallini, Artidoro Pagnoni, Tomasz Limisiewicz, Gargi Ghosh, Luke
  Zettlemoyer, Christopher Potts, Xiaochuang Han, and Srinivasan Iyer.
\newblock Fast {Byte Latent Transformer}.
\newblock \emph{arXiv preprint arXiv:2605.08044}, 2026.

\bibitem[Lin et~al.(2024)Lin, Gou, Gong, Liu, Shen, Xu, Lin, Yang, Jiao, Duan,
  and Chen]{lin2024rho1}
Zhenghao Lin, Zhibin Gou, Yeyun Gong, Xiao Liu, Yelong Shen, Ruochen Xu, Chen
  Lin, Yujiu Yang, Jian Jiao, Nan Duan, and Weizhu Chen.
\newblock Not all tokens are what you need for pretraining.
\newblock In \emph{Advances in Neural Information Processing Systems}, 2024.
\newblock \doi{10.52202/079017-0914}.
\newblock arXiv:2404.07965 (as Rho-1: Not All Tokens Are What You Need).

\bibitem[Liu et~al.(2026)Liu, Yu, Zhang, Gao, and Zhang]{liu2026entropymoe}
Bo~Liu, Muxuan Yu, Yu~Zhang, Pengfei Gao, and Yongping Zhang.
\newblock {EntropyMoE}: Entropy-aware sparse expert routing for tokenizer-free
  {LLMs}.
\newblock \emph{arXiv preprint arXiv:2608.06398}, 2026.

\bibitem[Meister(2026)]{meister2026tokeval}
Clara Meister.
\newblock {TokEval}: A tokenizer evaluation suite.
\newblock In \emph{Conference on Language Modeling (COLM)}, 2026.
\newblock arXiv:2608.18062.

\bibitem[Nawrot et~al.(2023)Nawrot, Chorowski, {\L}a{\'n}cucki, and
  Ponti]{nawrot2023dynamic}
Piotr Nawrot, Jan Chorowski, Adrian {\L}a{\'n}cucki, and Edoardo~M. Ponti.
\newblock Efficient transformers with dynamic token pooling.
\newblock In \emph{Proceedings of the 61st Annual Meeting of the Association
  for Computational Linguistics}, 2023.
\newblock arXiv:2211.09761.

\bibitem[Pagnoni et~al.(2025)]{pagnoni2025blt}
Artidoro Pagnoni et~al.
\newblock {Byte Latent Transformer}: Patches scale better than tokens.
\newblock In \emph{Proceedings of the 63rd Annual Meeting of the Association
  for Computational Linguistics (Long Papers)}, 2025.
\newblock arXiv:2412.09871. Code: https://github.com/facebookresearch/blt.

\bibitem[Paster et~al.(2024)Paster, Dos~Santos, Azerbayev, and
  Ba]{paster2024openwebmath}
Keiran Paster, Marco Dos~Santos, Zhangir Azerbayev, and Jimmy Ba.
\newblock {OpenWebMath}: An open dataset of high-quality mathematical web text.
\newblock In \emph{International Conference on Learning Representations
  (ICLR)}, 2024.
\newblock arXiv:2310.06786.

\bibitem[Raposo et~al.(2024)]{raposo2024mod}
David Raposo et~al.
\newblock {Mixture-of-Depths}: Dynamically allocating compute in
  transformer-based language models.
\newblock \emph{arXiv preprint arXiv:2404.02258}, 2024.

\bibitem[Singh and Strouse(2024)]{singh2024tokenization}
Aaditya~K. Singh and DJ~Strouse.
\newblock Tokenization counts: the impact of tokenization on arithmetic in
  frontier {LLMs}.
\newblock \emph{arXiv preprint arXiv:2402.14903}, 2024.

\bibitem[Slagle(2024)]{slagle2024spacebyte}
Kevin Slagle.
\newblock {SpaceByte}: Towards deleting tokenization from large language
  modeling.
\newblock In \emph{Advances in Neural Information Processing Systems}, 2024.
\newblock arXiv:2404.14408.

\bibitem[Vera~Z{\'u}{\~n}iga(2026)]{verazuniga2026patch}
Nicol{\'a}s Vera~Z{\'u}{\~n}iga.
\newblock When the next byte's type is fixed: why entropy patching skips
  computed values under tight budgets.
\newblock \emph{arXiv preprint arXiv:2610.22100}, 2026.
\newblock Version 2; version 1 was titled ``Patch where the text is new: a
  compression view of where byte-level models should spend compute''. Version 2
  corrects the main claims of version 1.

\bibitem[Videau et~al.(2025)Videau, Youbi~Idrissi, Leite, Schoenauer, Teytaud,
  and Lopez-Paz]{videau2025aunet}
Mathurin Videau, Badr Youbi~Idrissi, Alessandro Leite, Marc Schoenauer, Olivier
  Teytaud, and David Lopez-Paz.
\newblock From bytes to ideas: Language modeling with autoregressive {U-Nets}.
\newblock In \emph{Advances in Neural Information Processing Systems}, 2025.
\newblock arXiv:2506.14761.

\bibitem[Zheng et~al.(2026)Zheng, Bashlovkina, Dozat, Garrette, Rimell, and
  Maynez]{zheng2026scratchpad}
Lin Zheng, Vasilisa Bashlovkina, Timothy Dozat, Dan Garrette, Laura Rimell, and
  Joshua Maynez.
\newblock {Scratchpad Patching}: Decoupling compute from patch size in
  byte-level language models.
\newblock \emph{arXiv preprint arXiv:2605.09630}, 2026.

\end{thebibliography}

\section{Appendix A. Experimental
standards}\label{appendix-a.-experimental-standards}

The project ran as a partly autonomous research loop, and several early
conclusions were wrong. The standards in \texttt{program.md} exist
because breaking each produced a wrong result: a word-start rule that
read the byte it was predicting (caught by an automatic causality test,
\texttt{test\_causal.py}); effects at two seeds that were not detectable
at five (§12); a synthetic generator that placed dependencies outside
the model's context; an effect that was mostly architectural;
undertrained models hiding the effect; and a rule that exploited the
generator's fixed number of variable names. Trained-model comparisons
use three or more seeds except where a table or the text says otherwise:
in Table 3, 1.1M dependence and 53.7M hand-written have two seeds, and
jump and entropy at 20\% one; in the text, entropy plus dependence (§7)
and the window-32 trace models (§9) have two. BLT-1B comparisons use
paired tests, and every rule is checked for causality.

\end{document}